\documentclass[journal]{IEEEtran}

\usepackage[utf8]{inputenc} 
\usepackage[T1]{fontenc}    

\usepackage{amsmath,amsfonts}
\usepackage{algorithmic}
\usepackage{array}
\usepackage[caption=false,font=normalsize]{subfig}
\usepackage{textcomp}
\usepackage{stfloats}
\usepackage{url}
\usepackage{xurl}
\usepackage{verbatim}
\usepackage{graphicx}
\usepackage{balance}

\usepackage{colortbl}
\usepackage{tabulary}
\definecolor{Gray}{gray}{0.9}
\usepackage{amsmath}
\usepackage{subfloat}
\usepackage{hyperref}       
\usepackage{url}            
\usepackage{booktabs}       
\usepackage{amsfonts}       
\usepackage{nicefrac}       
\usepackage{microtype}      
\usepackage{xcolor}         
\usepackage{anyfontsize}
\usepackage{multirow}
\usepackage{wrapfig}
\usepackage[numbers,sort]{natbib}
\usepackage{enumitem}
\usepackage{amsmath,amsthm,amssymb}
\usepackage{algorithm}
\usepackage{algorithmic}
\usepackage{graphicx}
\usepackage{colortbl}
\usepackage{dsfont}
\usepackage{float} 
\usepackage{caption}
\usepackage{booktabs}
\usepackage{makecell}
\usepackage{diagbox}
\usepackage{fancybox}
\usepackage{bm}
\usepackage{color}

\hypersetup{
	colorlinks=true,
	linkcolor=black,
	filecolor=blue,      
	urlcolor=blue,
	citecolor=blue,
}

\newcommand{\partitle}[1]{\vspace{0.3em} \noindent \textbf{#1.}}
\def\ie{$i.e.$}
\def\eg{$e.g.$}

\begin{document}
\title{Rethinking Data Protection in the (Generative) Artificial Intelligence Era}

\author{%
\IEEEauthorblockN{%
\textbf{Yiming Li}$^{1,2}$,
\textbf{Shuo Shao}$^{1}$,
\textbf{Yu He}$^{1}$,
\textbf{Junfeng Guo}$^{3}$,
\textbf{Tianwei Zhang}$^{2}$,
\textbf{Zhan Qin}$^{1}$,
\textbf{Pin-Yu Chen}$^{4}$,\\
\textbf{Michael Backes}$^{5}$,
\textbf{Philip Torr}$^{6}$,
\textbf{Dacheng Tao}$^{2}$,
\textbf{Kui Ren}$^{1}$
}%
\\[1ex]
\IEEEauthorblockA{$^{1}$The State Key Laboratory of Blockchain and Data Security, Zhejiang University}\\
\IEEEauthorblockA{$^{2}$Nanyang Technological University}
\IEEEauthorblockA{$^{3}$University of Maryland}
\IEEEauthorblockA{$^{4}$IBM Research}\\
\IEEEauthorblockA{$^{5}$CISPA Helmholtz Center for Information Security}
\IEEEauthorblockA{$^{6}$University of Oxford}\\[0.5ex]
\IEEEauthorblockA{%
\href{mailto:liyiming.tech@gmail.com}{liyiming.tech@gmail.com}; \href{mailto:qinzhan@zju.edu.cn}{qinzhan@zju.edu.cn}
}
\thanks{The first two authors contributed equally to this paper. }
\thanks{This work was partly completed while Yiming Li was a Research Professor at Zhejiang University; he is now with Nanyang Technological University.}
}



\markboth{Perspective}%
{Perspective}

\maketitle

\begin{abstract}
The (generative) artificial intelligence (AI) era has profoundly reshaped the meaning and value of data. No longer confined to static content, data now permeates every stage of the AI lifecycle—from the training samples that shape model parameters to the prompts and outputs that drive real-world model deployment. This shift renders traditional notions of data protection insufficient, while the boundaries of what needs safeguarding remain poorly defined. Failing to safeguard data in AI systems can inflict societal and individual harm, underscoring the urgent need to clearly delineate the scope of and rigorously enforce data protection. In this perspective, we propose a four-level taxonomy, including non-usability, privacy-preservation, traceability, and deletability, that captures the diverse protection needs arising in modern (generative) AI models and systems. Our framework offers a structured understanding of the trade-offs between data utility and control, spanning the entire AI pipeline, including training datasets, model weights, system prompts, and AI-generated content. We analyze representative technical approaches at each level and reveal regulatory blind spots that leave critical assets exposed. By offering a structured lens to align future AI technologies and governance with trustworthy data practices, we highlight the urgent need to rethink data protection for modern AI techniques and provide timely guidance for developers, researchers, and regulators alike.

\end{abstract}


\section{Introduction}

Artificial Intelligence (AI) has experienced tremendous progress in the last few decades and is widely and successfully deployed in almost all domains, such as identity verification, e-commerce, and healthcare \cite{lecun2015deep, wang2023multitask, thirunavukarasu2023large,chen2025introduction}. With the recent rapid development of AI-enpowered generative models ($e.g.$, large language model (LLM) \cite{zhang2023survey} and diffusion model \cite{croitoru2023diffusion}), people can use them to easily generate high-quality images, audio, video, and text (instead of simple predictions). More importantly, these powerful models are close at hand, where users can simply exploit them via APIs ($e.g.$, GPT-4 \cite{achiam2023gpt} and Midjourney \cite{midjourney}) or even directly download them from open-source communities/platforms ($e.g.$, Hugging Face). Arguably, we have moved into the era of (generative) AI.

\begin{figure}[!t]
    \centering
    \includegraphics[width=\linewidth]{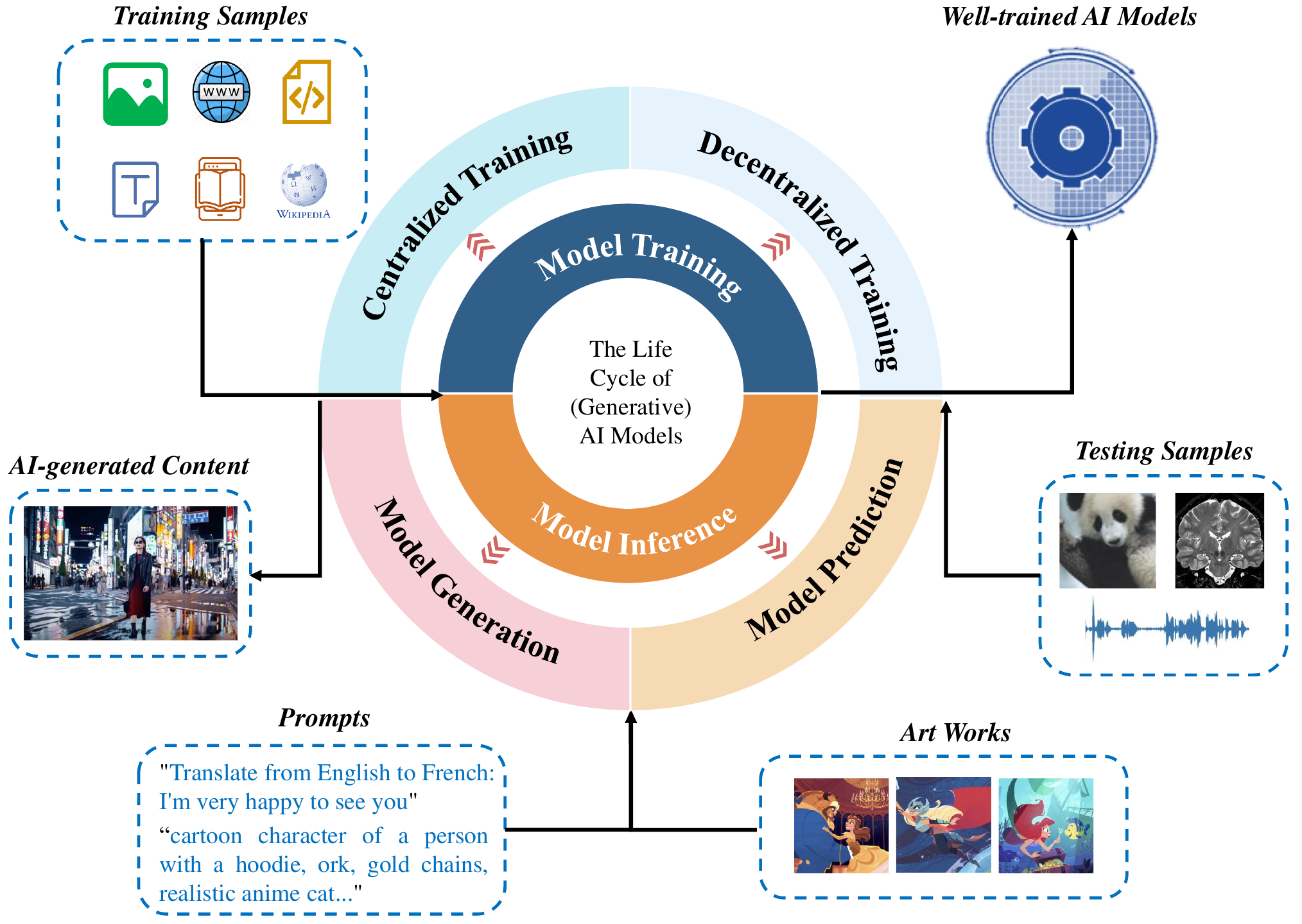}
    \vspace{-1.5em}
    \caption{Data flow across the life‑cycle of a (generative) AI model. The schematic traces how different forms of data emerge and circulate from the moment raw samples are collected to the point at which a deployed model generates new content. \textbf{(i) Data Collection and Curation}:  Samples, such as images, texts, and audio clips are gathered and annotated; once aggregated, they form the training dataset that drives model learning and the testing dataset used for validation. \textbf{(ii) Model Training}: These datasets are transformed into model parameters ($e.g.$, weights and biases), turning the well‑trained model itself into a valuable, model‑centric data asset. \textbf{(iii) Model Inference}: After deployment, users supply inputs or prompts—which may contain private or proprietary information—that the model processes to produce AI‑generated content ranging from class labels to code, images, or full documents. Arrows indicate how each artefact ($e.g.$, dataset, model parameters, prompts, and outputs) can be independently copied, released, or shared, underscoring why all of them must be considered within a comprehensive data‑protection framework.}
    \label{fig:intro}
    \vspace{-2em}
\end{figure}

In general, the prosperity of AI heavily relies on high-quality data, with which researchers and developers can train, evaluate, and improve their models. For example, advanced LLMs like GPT-4~\cite{achiam2023gpt} and DeepSeek~\cite{guo2025deepseek} required vast, curated datasets from diverse sources, often refined with costly human feedback to ensure quality and alignment. Similarly, specialized medical models like Google's Med-PaLM~\cite{Med-PaLM}, designed for clinical question answering and summarization, or diagnostic AI systems for tasks like cancer detection from images, relied heavily on large, diverse clinical datasets ($e.g.$, the Cancer Genome Atlas (TCGA)~\cite{tcga}) meticulously annotated by medical experts, a complex and resource-intensive necessity.
In particular, collecting and annotating data remains a significant obstacle for most companies since they are time-consuming and even expensive~\cite{liang2022advances}. Accordingly, these data are undoubtedly valuable assets to their owners and deserve to be protected.

Data protection has long been a critical area of research due to its significance in safeguarding the legitimate rights of data owners. Various regulations, such as GDPR~\cite{GDPR2016} or EU AI Act~\cite{EUAIAct2024}, highlighted the importance of data protection. In the past, data typically existed as discrete digital items, whose value was derived largely from their content. For example, it could be digitized artwork, photographs, videos, etc. Accordingly, traditional data protection mainly refers to protecting the content of data from unauthorized use and redistribution, although its specific definition and scope still remain ambiguous to some extent. In practice, data owners would encrypt files~\cite{boneh2001identity, martins2017survey, deng2020identity} before storage or transmission and embed digital watermarks~\cite{hartung1999multimedia, guan2022deepmih, ranjbar2023nft} when releasing data publicly or in digital marketplaces in the past.

However, in the AI era, especially with the emergence of generative AI models, the scope of data protection has become far more complex and ambiguous~\cite{rigaki2023survey, ziller2024reconciling,mittal2024responsible}. As shown in Figure~\ref{fig:intro}, data permeates every stage of an AI model's life cycle, making its value increasingly tied to the model rather than just the raw content of the data. For instance, developers compile many individual samples into large training datasets that feed into model development. The trained models themselves then become valuable data assets with significant commercial value. In addition, high-value or sensitive data ($e.g.$, original artworks or personal medical records) may also be incorporated as inputs during a model's inference stage. Besides, with the rise of generative AI models, the outputs of inference are no longer simple predictions – they can be substantial content in their own right. For example, an LLM might generate executable code for a requested function~\cite{he2025benchmarking}, or a diffusion model might produce a realistic image for an advertisement or animation clips~\cite{ho2020denoising}. These AI-generated outputs are also valuable forms of data that merit protection.

This ambiguity in scope makes meaningful protection and regulation difficult. For instance, in 2023, Samsung Electronics discovered that employees had inadvertently leaked proprietary source code by inputting it into OpenAI's ChatGPT, prompting it to prevent its staff from using such external generative AI tools on company systems~\cite{samsungban}; that same year, Italy's Data Protection Authority (Garante) imposed a nationwide suspension of ChatGPT after a leak of user conversations and allegations that personal data had been ingested for training without a lawful basis~\cite{italyban}. These incidents underscore the urgent need for a systematic understanding of what, precisely, must be protected against the backdrop of blooming AI-integrated applications and data markets. 

To tackle this problem, this paper offers the first timely overview and categorization of data protection in the (generative) AI era. Specifically, we introduce a hierarchical taxonomy of data protection comprising four distinct levels: data non-usability, privacy-preservation, traceability, and deletability. Each level in this taxonomy reflects a different balance between how usable the data remains for AI models and the degree of control or protection imposed on that data. At the most stringent end of the spectrum, data non-usability ensures that certain data cannot be used for model training or inference at all, offering maximal protection by completely sacrificing utility. Progressing down the hierarchy, privacy-preservation permits data to be utilized in model development and application while safeguarding sensitive information, a trade-off that maintains some utility but enforces confidentiality of personal or private attributes. Further along, traceability allows nearly full data usage, yet embeds mechanisms to track the data's origin and usage, thereby enabling transparency and accountability (for instance, detecting if data has been misused) with only minimal impact on the data's functionality. Finally, at the most permissive level, data deletability lets data be fully integrated on the condition that its influence can be later removed from the model upon request. This last level emphasizes post-hoc control (aligning with `right to be forgotten' principles) without impeding initial data utility. In particular, to ground this taxonomy, we systematically review representative technical approaches at each level, highlighting their strengths and limitations in practical settings.

By clearly delineating these four levels, our framework brings much-needed clarity to the often conflated notion of `data protection' in the (generative) AI era. Researchers and practitioners can now specify whether they aim to prevent any use of certain data, protect privacy during use, ensure traceable usage, or enable later deletion. This structured hierarchy not only highlights the progressive relaxation of restrictions (from strict non-use to full use with after-the-fact removal) but also helps disambiguate the scope of data protective measures in the AI era. Moreover, it provides a structured lens to evaluate existing legal and regulatory instruments: in the later section, we will show how existing national and international policies or regulations align (or fail to align) with each data protection level, illuminating where governance already supports these protective goals and where further action is required.

\section{Hierarchical Taxonomy of Data Protection}

\subsection{What Data Do We Need to Protect in the AI Era?}

In the (generative) AI era, the scope of data protection has expanded significantly, moving far beyond the traditional focus on static data content. Specifically, AI models generate and consume various forms of data throughout their lifecycle, from initial training to final inference. At each stage, different categories of data emerge as assets that warrant protection, whether for reasons of privacy, intellectual property, security, or commercial value. As presented in the previous section, Figure \ref{fig:intro} illustrates this lifecycle, where raw samples become training datasets, which in turn yield models; those models are then deployed to handle user prompts and produce AI-generated outputs. Every artifact along this chain, such as the training datasets, the trained model, the user inputs/prompts, and the AI-generated content (AIGC), carries its own significance and sensitivities. Below, we examine why each of these data categories matters and why they must be safeguarded within a comprehensive protection framework.

\vspace{0.3em}
\textbf{Training Datasets}: In the development phase of a model, large curated datasets serve as the fuel for learning. These collections of samples (images, text, audio, \emph{etc}.) are often aggregated from diverse sources, which inherently raises the risk of including sensitive personal information or copyrighted material \cite{longpre2025bridging}. Protecting training data is therefore crucial for legal and ethical reasons: developers must respect privacy rights and intellectual property (\eg, avoiding unauthorized use of personal photos~\cite{abadi2016deep} or scraping of copyrighted text~\cite{li2023black}) to comply with regulations and moral norms. Moreover, assembling and labeling high-quality datasets is expensive and time-consuming, making them commercially valuable assets for the organizations that curate them. Companies treat their data as proprietary know-how. For example, the success of ImageNet~\cite{deng2009imagenet} spurred competitive advantages in computer vision and beyond~\cite{lecun2015deep}. If such a dataset were stolen or misused, the original collector could suffer a significant loss of investment and competitive edge. For all these reasons, training data merits strong protection. This includes measures to preserve privacy (\eg, removing or anonymizing personal identifiers~\cite{sweeney2002k}) and to enforce rights management, ensuring the data is not redistributed or used beyond its permitted scope~\cite{huang2021unlearnable, li2023black, du2025sok}. In some cases, dataset owners even embed subtle markers (\eg, watermarks or fingerprints) into the data to enable traceability \cite{li2022untargeted, guo2024zero, li2025towards, zhu2024duwak,zhu2025tabwak}, so that if the data appears in an unauthorized model or repository, it can be identified and linked back to the source. Overall, securing the training dataset is the first pillar of data protection in the AI pipeline, preventing downstream issues that could arise from contaminated or compromised training information.

\vspace{0.3em}
\textbf{Trained Models}: Once an AI model has been trained, the model itself, encompassing its architectural configuration and learned parameters, becomes a model-centric data asset of immense value. Unlike raw training datasets, a trained model encapsulates generalizations drawn from potentially vast training data~\cite{croitoru2023diffusion,yuksekgonul2025optimizing,li2025move}. In effect, it is a compressed repository of that data's information. This gives the model significant commercial and strategic significance. Organizations invest heavily in developing high-performing models, and the resulting structure and weights are often regarded as trade secrets or key intellectual property. For example, the parameters of a state-of-the-art language model or image recognition network can confer a competitive edge, making the model file itself as sensitive as any proprietary dataset. Protecting this trained model data is therefore paramount – if it is exposed or stolen, an adversary or competitor could reuse it, undermining the original owner's investment and advantage~\cite{oliynyk2023know, shao2025explanation,wang2025sleepermark}. Accordingly, the trained model must be safeguarded much like any confidential dataset in the AI era, especially to preserve the commercial integrity of the model as a proprietary asset.

\vspace{0.3em}
\textbf{Deployment‑integrated Data}: Beyond the model's learned parameters, modern AI deployments usually incorporate additional auxiliary data that plays a crucial role in shaping their inference performance. These data are introduced at the deployment or runtime stage (after model training), and while not part of the model's weights, they effectively become extensions of the model's knowledge and policy. Two prominent examples are \emph{system prompts}~\cite{wu2024leveraging, polak2024extracting} and \emph{external knowledge bases}~\cite{lewis2020retrieval, salemi2024evaluating} used in conversational AI and retrieval-augmented generation (RAG). Such deployment-integrated data elements are often invisible to end-users but are pivotal in determining how the model responds to inputs. Importantly, they may embed sensitive or proprietary information, and their compromise can be just as damaging as a leak of the model itself. Even though this data is not `learned' during training, it must be protected because it directly influences the model's outputs and can inadvertently reveal protected information if misused. Specifically, system prompts are predefined directives or contexts given to a model at inference time, especially in large language model (LLM) deployments. For instance, a ChatGPT-like assistant might have a hidden prompt saying: `You are an expert medical assistant. Always answer with evidence-based information and in a reassuring tone.' This prompt is not part of the model's parameters but is provided by the developers to guide the model's behavior and set boundaries on its responses. System prompts help ensure consistency, align the model with ethical or style guidelines, and can embed institutional knowledge and policies, or even achieve differentiated services through carefully designed prompts. Because they often encode rules and content that the provider considers sensitive (including possibly proprietary instructions or content examples), system prompts are sensitive deployment data~\cite{yao2024promptcare, shen2024prompt}. If an adversary were to discover the exact contents of these prompts, they might exploit them (\eg, by crafting inputs that override and manipulate the system instructions, or developing competitive applications by illegally acquiring the system prompts). External knowledge bases are specialized repositories of curated information, integrated at inference time to enhance the capability of AI models (especially LLMs) through a mechanism known as RAG. Unlike system prompts, external knowledge bases are extensive collections of documents or structured data that models dynamically retrieve and incorporate into their reasoning process to produce accurate, timely, and domain-specific responses. For example, medical assistants powered by retrieval-augmented large language models (RA-LLMs) might access confidential diagnostic records to inform clinical decisions, while financial agents leverage internal market databases for precise forecasting. Although external knowledge bases are not part of the trained model parameters, their content may be highly sensitive, often comprising proprietary or confidential information crucial to an organization's operational advantage~\cite{fan2024survey}. Together, these examples highlight that deployment-integrated data, exemplified by system prompts and external knowledge bases, represent critical yet often overlooked data assets whose protection is also indispensable in today's (generative) AI era.

\vspace{0.3em}
\textbf{User's Input}: When a model is deployed, new data enters the picture: the inputs (especially prompts) supplied by users during inference. These inputs can be as trivial as a search query or as sensitive as a detailed medical history or proprietary source code, depending on the application~\cite{liu2021machine, rigaki2023survey}. In the AI era, particularly with the rise of accessible generative AI chatbots and assistants, users routinely provide personal or confidential data to AI systems in exchange for tailored outputs. It is imperative to protect this prompt data for privacy, security, and ethical reasons. From a privacy standpoint, any personal information in a user's query (names, addresses, health details, etc.) should be handled in compliance with data protection laws and the user's expectations of privacy. There have already been real-world incidents underscoring this need: for example, in 2023, Italy temporarily banned ChatGPT over concerns that the platform was not adequately protecting user-provided personal data~\cite{italyban}. Commercial confidentiality is equally at stake – consider an employee who uses an AI coding assistant and enters proprietary code as a prompt. If the AI service retains this input, it could lead to an unintended leak of trade secrets. This scenario is not hypothetical: employees at Samsung accidentally disclosed confidential source code and meeting notes by submitting them to ChatGPT, which retained those prompts on its servers~\cite{samsungban}. To address such issues, techniques like robust access control~\cite{yu2010achieving, han2021blockchain} and privacy guarantees~\cite{huang2022cheetah, zhang2025secure} must be in place at the inference stage. Ethically, users should have transparency and agency regarding their inputs – they should know if prompts will be logged or used for training, and ideally have the right to deletion (aligning with the `right to be forgotten' in privacy regulations). Protecting users' input data not only complies with privacy laws but also builds trust. If users fear their prompts might be misused or leaked, they will be reluctant to adopt AI solutions, limiting the technology's benefits. Thus, safeguarding users' input is now a fundamental component of data protection in AI, aimed at preserving individual privacy and maintaining confidentiality in AI services \cite{qu2025prompt}.

\vspace{0.3em}
\textbf{AI-generated Content (AIGC)}: The final category of protected data arises from the model's own outputs. In particular, instead of simple numbers, modern (generative) AI systems can produce rich content like paragraphs of text, realistic images, and code snippet~\cite{driess2023palm, haase2025towards}. These AIGCs have already become valuable digital objects~\cite{ren2024sok, zhao2025sok}. While the standalone content of AIGC has inherent protection needs related to intellectual property, ownership, and potential sensitivities~\cite{samuelson2023generative,hu2023radar,he2024rigid,li2025we}, our primary focus here aligns with the model-centric perspective: protecting AIGC in its role as a data asset within the (generative) AI ecosystem. Given its high fidelity and utility, AIGC is increasingly leveraged not just as a final product, but also as \textit{data} that feeds back into the AI cycle. For example, AIGC is valuable for creating large-scale synthetic datasets, for knowledge distillation~\cite{gou2021knowledge}, or as deployment-integrated data (\eg, instances used in retrieval-augmented generation). Protecting AIGC in this capacity is therefore crucial. This can involve ensuring traceability to understand its provenance if used for training~\cite{li2025towards}, or employing mechanisms akin to non-usability or access control to prevent unauthorized reuse for training competing models. Our framework thus emphasizes the governance needed when this generated content itself becomes data for subsequent model training or inference, highlighting its flow within the broader (generative) AI model's lifecycle.

\vspace{0.3em}
In conclusion, data protection in the (generative) AI era must extend across the model's entire lifecycle. From the raw training dataset, to the trained model, to the prompts it processes and the content it generates, each component contains information that could be sensitive, proprietary, or otherwise regulated. Notably, each type of data can be copied or transmitted independently – one can leak a dataset, steal a model's weights, expose a user's prompt, or misappropriate an AI output, which is why all of them must be considered in a holistic protection strategy. By clearly identifying these categories, we can align specific protection goals and techniques to each: \eg, privacy-preservation for personal data in training sets and prompts, traceability mechanisms for outputs, and so forth. The following sections will build on this lifecycle view to explore how a hierarchical taxonomy can collectively safeguard the myriad data assets in the AI era, and how emerging data protection techniques map onto each protection level.

\begin{figure*}[!t]
    \centering
    \vspace{-1em}
    \includegraphics[width=0.95\linewidth]{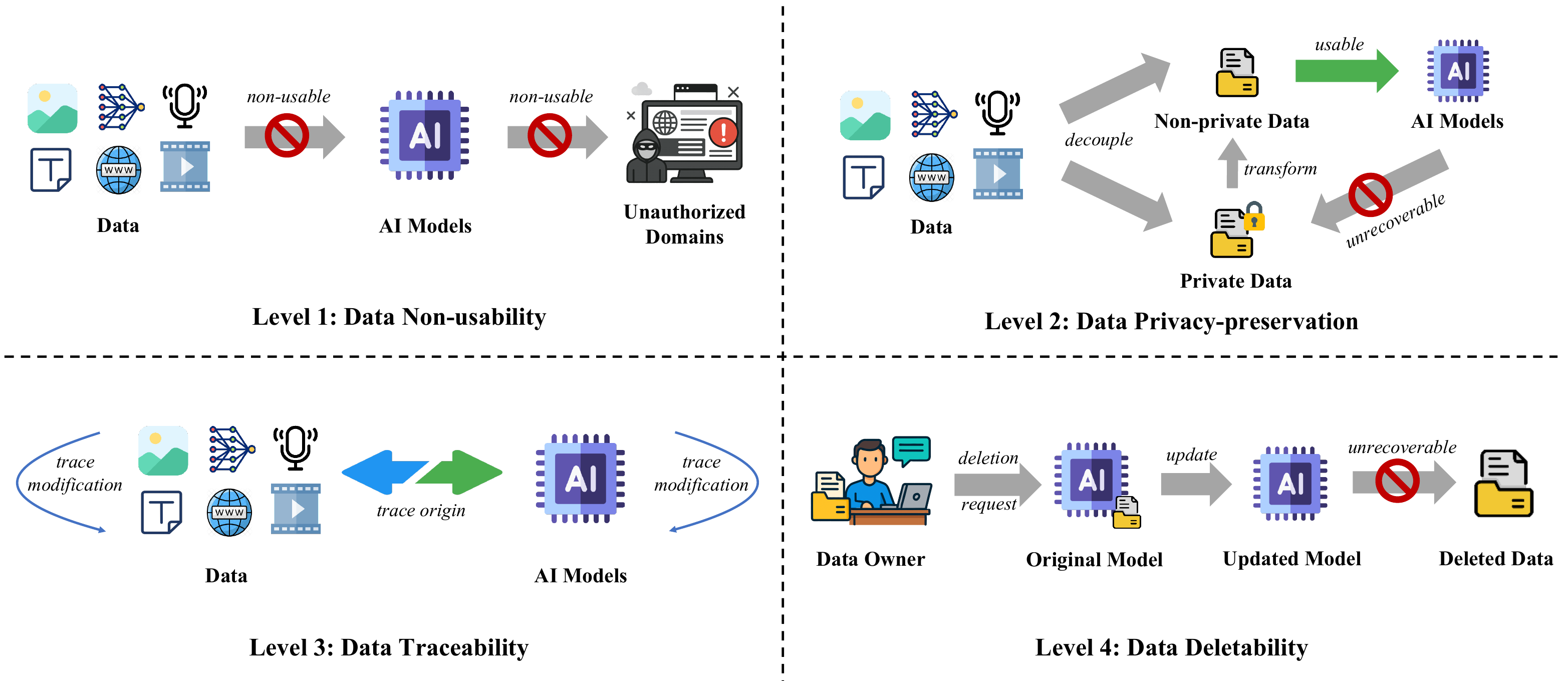}
    \vspace{-0.3em}
    \caption{Hierarchical taxonomy of data protection in the (generative) AI era. This taxonomy comprises four distinct protection levels, each representing a trade-off between data usability and the degree of protection provided. At the most stringent level, \textbf{data non-usability} completely restricts the use of specific data in model training and inference, thus offering maximal protection at the cost of total data utility. The next level, \textbf{data privacy-preservation}, allows data use under stringent privacy safeguards, enabling some practical utility while protecting sensitive or private attributes. Moving further, \textbf{data traceability} permits extensive data usage but integrates methods to track data origins and modifications, supporting transparency and accountability with minimal functional interference. At the most permissive level, \textbf{data deletability} places no initial restriction on data usage but mandates mechanisms for fully removing data's influence from trained models post hoc, aligned with principles such as the `right to be forgotten'. This hierarchical taxonomy helps disambiguate the scope of data protective measures and provides a structured lens to evaluate and further design related regulations in protecting data in the (generative) AI era.}
    \label{fig:hierarchy}
    \vspace{-1em}
\end{figure*}

\subsection{Towards the Hierarchical Taxonomy of Data Protection}

\textbf{Taxonomy Overview.} 
AI's data-protection challenges span a spectrum from extremely strict control of data to more permissive use with after-the-fact safeguards. To make sense of this spectrum, we propose a four-level hierarchical taxonomy of data protection: data non-usability, data privacy-preservation, data traceability, and data deletability. Each successive level in this hierarchy relaxes the protections on data slightly, trading off some degree of control for greater data utility. At the highest, the most restrictive end, data non-usability ensures that certain data cannot be used for model training or inference at all, thereby offering maximum protection by completely sacrificing that data's utility. Stepping down one level, data privacy-preservation permits data to be employed in model development or inference while safeguarding sensitive information – a compromise that preserves some utility but enforces confidentiality of personal or private attributes. Next, data traceability allows nearly full use of data for AI models, yet embeds mechanisms to track the data's origin, usage, and modifications (\eg, to detect if data has been misused), thereby enabling transparency and accountability with only minimal impact on the data's functionality. Finally, at the most permissive level, data deletability imposes nearly no restriction on a dataset's initial use for training and inference, instead requiring that the data's influence can later be removed from the model upon the user's requests. This last level emphasizes post-hoc control (aligning with `right to be forgotten' principles) without impeding the data's immediate usefulness. Figure~\ref{fig:hierarchy} illustrates this hierarchy of protection levels, which forms a clear gradient from strong protection/low utility at Level 1 to low protection/high utility at Level 4.

\vspace{0.3em}
\textbf{Level 1: Data Non-usability. } 
Data non-usability encompasses methods that intentionally render certain data entirely useless for AI applications, including training and inference, even if that data is publicly available. In essence, it ensures that specified data cannot contribute to model learning or predicting whatsoever. This is crucial in scenarios where individuals or organizations demand strict control over how their data is utilized by AI systems. For instance, authors and journalists have voiced objections to their articles or books being used to train language models without consent~\cite{abdali2024decoding}; similarly, visual artists often share their works online but may strongly oppose using AI models to transfer their style to others during inference~\cite{shan2023glaze}. By completely precluding any use of the data in model development, data non-usability offers the most stringent level of protection in our taxonomy – achieving maximal data control at the expense of all potential utility.

\vspace{0.3em}
\textbf{Level 2: Data Privacy-preservation.} Data Privacy-Preservation focuses on protecting sensitive information within data while still allowing the data to be used for developing AI models or producing meaningful responses/inferences~\cite{dwork2006calibrating, li2024safeear}. This approach is especially critical in sectors like healthcare, social media, and online services—domains where large volumes of personal data (\eg, age, gender, location, or purchasing behavior) are routinely collected and analyzed~\cite{meurisch2021data}. For instance, a hospital or research institute might analyze patient records to train a disease-detection model, but it must do so without exposing any individual's identity or private details. Users also do not want to leak their private information when chatting with AI chatbots interacting with prompts. Ensuring privacy is not only a legal obligation for data handlers, but also a crucial measure to prevent misuse of personal information and to maintain public trust in AI-driven technologies and applications. In practice, privacy-preserving measures mean that a significant portion of each dataset (namely, the privacy-sensitive attributes) is withheld, masked, or otherwise not directly accessible during training or inference~\cite{shokri2015privacy}. Consequently, data privacy-preservation still represents a high level of protection for the data, second only to complete non-usability in its restrictiveness, while enabling much more data utility than the latter.

\vspace{0.3em}
\textbf{Level 3: Data Traceability.} Data Traceability refers to the ability to track the origin, history, and influence of data as it is used in AI applications during training and inference. This capability allows stakeholders to audit and verify data usage. For example, an individual might want to check whether their personal data was incorporated into a model for training or generating works of art without permission, and a model developer might need to detect if a training dataset or a pre-trained model has been tampered with or misused and avoid the potential backdoor in them~\cite{li2022backdoor, alber2025medical, chen2025refine}. By enabling such oversight, traceability measures greatly enhance transparency and accountability in how data fuels AI systems. Importantly, implementing traceability need not significantly hinder the data's usefulness for modeling: the data remains almost fully available for training or inference, with at most slight modifications introduced to embed identifiers (\eg, imperceptible watermarks or metadata tags) that enable later tracking~\cite{liu2024model,li2025towards}. Thus, data traceability provides a more moderate level of protection – less restrictive than privacy-preservation since it leaves the data content largely intact, but still offers an important safeguard through post hoc auditability.

\vspace{0.3em}
\textbf{Level 4: Data Deletability.} Data deletability is the capacity to completely remove a specific piece of data and its influence from a trained (AI) model. While deleting a data file from a storage database is trivial, eliminating that data's imprint on an AI model is a far more challenging task~\cite{bourtoule2021machine}. This level of protection ensures that if a particular data sample must be purged – for example, because it is no longer needed or because the individual who provided the data withdraws consent – there is a mechanism to do so cleanly and effectively. Such capability is particularly pertinent to user rights and data governance frameworks (\eg, complying with the `right to be forgotten' in GDPR regulations~\cite{GDPR2016}). Notably, enabling deletability does not require compromising the data's utility during initial model training; the data can be used to its full extent upfront, and the protective measure comes into play only later, if and when deletion is required. Because this approach imposes no upfront usage restrictions, it offers the lowest immediate level of protection among the four levels – instead, its strength lies in allowing retrospective removal. In summary, data deletability prioritizes giving data owners ultimate control after model development, even though it provides only minimal protection at the time of data use.

\section{Techniques for Data Protection}

\begin{figure*}[!t]
    \centering
    \vspace{-1.5em}
    \includegraphics[width=0.98\linewidth]{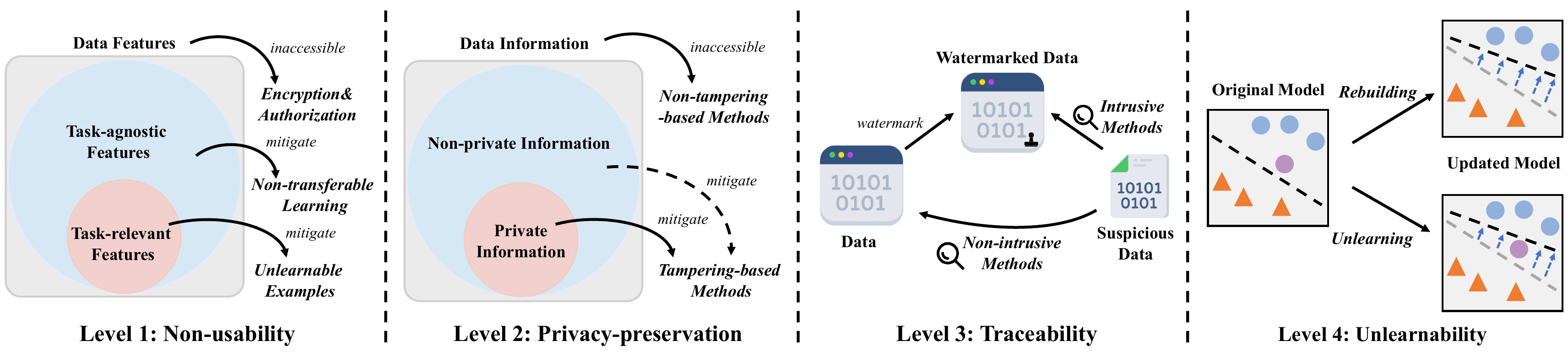}
    \caption{Design principles of techniques for each level. \textbf{Level 1. Non-usability}: Encryption and (fine-grained) authorization confine direct data access solely to authorized parties, while techniques such as unlearnable examples and non-transferable learning disable data exploitation in unauthorized domains by mitigating particular data features, thereby achieving non-usability indirectly; \textbf{Level 2. Privacy-preservation}: These techniques generally fall into two main categories: tampering-based and non-tampering-based methods. The former perturbs private portions of the data (occasionally at the cost of tampering with some non-private content), whereas the latter prevents direct access without data modification while preserving data utilities; \textbf{Level 3. Traceability}: Traceability techniques intrusively attach ownership signals (\ie, watermarks) to original data or directly infer provenance and potential modifications non-intrusively by analyzing data's intrinsic information; \textbf{Level 4. Deletability}: The influence of protected data (denoted by `purple circle' in the sub-figure) can be removed either by excising the data and rebuilding the AI model from scratch to directly change the decision surface (marked in `black dot-line') or, more efficiently, by targeted unlearning that erases its influence (to the surface) without full model reconstruction, thereby ensuring data deletability.}
    \label{fig:technique}
    \vspace{-1em}
\end{figure*}

To translate the conceptual taxonomy of data protection into practice, this section briefly describes a range of design principles and corresponding representative techniques tailored to the four protection levels introduced above. Figure \ref{fig:technique} illustrates the design principles of techniques for each level.

\partitle{Techniques for Non-usability} Non-usability encompasses strategies that block any unauthorized party from using or even accessing protected data. Arguably, the most direct method is encryption \cite{shannon1949communication,diffie1976new,purcell2018encryption}: by securing data with strong cryptographic keys, the information remains unintelligible without proper authorization. A complementary line of defense ensures that the data cannot be exploited even if an adversary obtains it. For example, authorization mechanisms, including fine-grained data-access control \cite{yu2010achieving,yang2013dac,han2021blockchain} and model-level authorization \cite{ren2022protecting,ren2024active}, allow only approved entities to obtain (correct) model outputs. Unauthorized requests receive degraded or nonsensical responses. Beyond controlling access or general utility, a further class of techniques makes the data unusable in unauthorized domains: unlearnable examples \cite{huang2021unlearnable,shan2023glaze} embed imperceptible perturbations that frustrate a model's ability to extract `task-relevant' features, whereas non-transferable learning \cite{wang2021non,hong2024improving} deliberately suppresses `task-agnostic' features so that any knowledge gleaned cannot generalize to unintended tasks. Together, these techniques align with a `secure-by-design' philosophy: the data would be essentially non-usable and remain protected even in worst-case scenarios.

\partitle{Techniques for Privacy-preservation} Privacy-preservation techniques enable the beneficial use of data for AI development while shielding sensitive information. They fall into tampering-based and non-tampering-based categories. In tampering-based approaches, the data themselves (at least their private components) are modified so that private/sensitive attributes become indistinguishable or masked  \cite{gong2023gan}. For example, early schemes such like k-anonymity and L-diversity schemes generalize or suppress identifying details \cite{sweeney2002k,machanavajjhala2007diversity}, though they can reduce utility and remain vulnerable to linkage attacks \cite{ganta2008composition}; Differential privacy \cite{dwork2006calibrating,utpala2023locally} provides a stronger guarantee by injecting carefully calibrated noise into data, intermediate computations, or outputs \cite{abadi2016deep,li2024local}. The added randomness masks each individual's contribution while remaining versatile enough for model training, synthetic-data generation, and attack mitigation \cite{jordon2018pate,ziller2024reconciling}. In contrast, non-tampering-based techniques avoid modifying raw data, seeking privacy protection with minimal impact on data utility. For example, homomorphic encryption \cite{paillier1999public,martins2017survey} enables computations on encrypted inputs, eliminating exposure during processing; Privacy-preserving distributed learning, such as federated learning \cite{mcmahan2017communication} and split learning \cite{gupta2018distributed}, keeps data local while sharing only aggregated model updates \cite{mcmahan2017communication,hanser2025data}. In this way, global models benefit from diverse datasets without centralizing sensitive records.

\partitle{Techniques for Traceability}
Traceability seeks to record and verify where data (including models) originate, how they are used, and whether they have been altered. Existing approaches can be broadly categorized into intrusive and non-intrusive methods. Intrusive methods embed explicit and external identifiers (dubbed `watermarks') into the data asset~\cite{guo2024domain, wei2024pointncbw}. For example, digital watermarking adds hidden yet robust signatures to datasets \cite{li2022untargeted,li2023black}, model parameters \cite{li2022defending,shao2024fedtracker}, or prompts \cite{yao2024promptcare}. These robust watermarks survive ordinary use and prove ownership. Contrarily, fragile watermarks \cite{zhang2008fragile,botta2021neunac} are deliberately brittle, breaking if tampered with, and therefore alerting potential modification. Non-intrusive methods, on the other hand, enable traceability by analyzing its intrinsic information or detecting its modifications without altering the underlying data asset. For example, membership inference \cite{shokri2017membership,he2025labelonly} evaluates whether a data point was in a model's training set; model fingerprinting \cite{cao2021ipguard,pasquini2025llmmap} probes a model with crafted inputs to reveal its identity; Cryptographic hashing \cite{merkle1987digital,rivest1991md4} produces unique fingerprints that change upon any bit-level alteration, while blockchain ledgers \cite{de2020survey,guo2021smartphone} maintain an immutable, time-stamped record of data states, making secret edits computationally infeasible.

\partitle{Techniques for Deletability}
Ensuring that specific data and their influence can be removed from AI models underpins rights such as GDPR's `right to be forgotten'. The most straightforward, yet costly, route is to directly delete the data and rebuild the AI model from scratch \cite{bourtoule2021machine,hu2024eraser}. A more efficient alternative is offered by unlearning techniques that specifically focus on erasing the \emph{influence} of the data instead of directly the content. These algorithms aim to approximate the model state that would have arisen had the targeted data never been used, thereby avoiding the significant expense and time required for complete retraining or rebuilding~\cite{guo2020certified, liu2023certified, jia2024wagle}.

\begin{table*}[!t]
\renewcommand{\arraystretch}{1.45}
    \centering
    \caption{The representative regulations of data protection in the (generative) AI era. The last column shows the levels of data protection covered in the regulation (\textbf{N}: non-usability, \textbf{P}: privacy-preservation, \textbf{T}: traceability, \textbf{D}: deletability).}
    \label{tab:regulation}
    \vspace{-0.5em}
    \scalebox{1.1}{
    \begin{tabular}{llll}
    \toprule
        Country/Region & Regulation Name & Year & Protection Level(s) \\
        \hline
        \multirow{2}{*}{USA} & California Consumer Privacy Act \cite{CCPA2018} & 2018 & \textbf{N, P, T, D}\\
        & Federal Zero Trust Data Security Guide \cite{federal2024} & 2024 & \textbf{N, P, T}\\
        \hline
        \multirow{4}{*}{EU} & General Data Protection Regulation \cite{GDPR2016} & 2016 & \textbf{N, P, T, D}\\
        & Ethics Guidelines for Trustworthy AI \cite{EthicsAIEU2019} & 2019 & \textbf{N, P, T}\\
        & EU AI Act \cite{EUAIAct2024} & 2024 & \textbf{N, P, T, D}\\
        & General-Purpose AI Code of Practice (Draft) \cite{GPAICodeDraftEU2025} & 2025 & \textbf{N, P, T}\\
        \hline
        \multirow{8}{*}{China} & Cybersecurity Law of the PRC \cite{CSLPRC2016} & 2016 & \textbf{N, P, T}\\
        & Data Security Law of the PRC \cite{DataSecurityLawPRC2021} & 2021 & \textbf{N, P, T}\\
        & Personal Information Protection Law of the PRC \cite{PIPLPRC2021} & 2021 & \textbf{N, P, T}\\
        & Administrative Measures for Generative Artificial Intelligence Services \cite{GenAIMeasuresPRC2023} & 2023 & \textbf{N, P, T}\\
        & Action Plan of the Development of Trustworthy Data Space \cite{TrustworthyDataSpacePRC2024} & 2024 & \textbf{N, P, T}\\
        & \makecell[l]{Implementation Plan on Improving Data Circulation Security Governance to Better \\ Promote the Marketization and Valorization of Data Elements \cite{DataCirculationPRC2025}} & 2025 & \textbf{N, P, T}\\
        & \makecell[l]{Methods for Identifying Synthetic Content
        Generated by Artificial Intelligence \cite{IdentifyAIGC2025}} & 2025 & \textbf{T}\\
        \hline
        \multirow{6}{*}{Others} & Artificial Intelligence Mission Austria 2030 \cite{AIMAT2030} & 2019 & \textbf{N, P, T}\\
        & Artificial Intelligence and Data Act \cite{AIDA2022} & 2022 & \textbf{P, T}\\
        & Brazilian AI Regulation \cite{BrazilAIRegulation2023} & 2023 & \textbf{N, P, T}\\
        & Enhancing Access to and Sharing of Data in the Age of Artificial Intelligence \cite{oecd2025enhancing} & 2025 & \textbf{N, P}\\
        & \makecell[l]{Joint Statement on Building Trustworthy Data Governance Frameworks\\ to Encourage Development of Innovative and Privacy-Protective AI \cite{JointStatementDataGov2025}} & 2025 & \textbf{N, P, T}\\
    \bottomrule
    \end{tabular}
    }
    \label{tab:regulations}
\end{table*}

\section{Regulations on Data Protection}

In the era of (generative) AI, regulation plays a foundational role in safeguarding data integrity, privacy, and accountability. Unlike traditional data systems, where protection focuses on static storage and access control, AI systems rely on dynamic, model-centric data use: once data is absorbed into a model's parameters, it may persist, influence downstream outputs, and defy straightforward removal. Legal frameworks thus serve as critical instruments to constrain unauthorized use, enforce privacy-preservation, ensure traceability, and empower users with redress and deletion rights. As shown in Table \ref{tab:regulations}, there are already some pioneering related regulations. For instance, many privacy laws operationalize L1 (Non-usability) by prohibiting the use of sensitive or unlawfully collected data for AI training altogether. Similarly, L2 (Privacy-preservation) is widely mandated through consent requirements, anonymization, and processing limits. Emerging regulations now also touch on L3 (Traceability)—requiring documentation of data provenance and logging of model operations—and even aspire to L4 (Deletability), allowing individuals to remove their data's influence post-training. As the diffusion of data across AI pipelines complicates direct user control, the regulation remains the strongest binding force for aligning model development with ethical and societal norms.

Globally, several regulatory regimes have responded to this challenge with varying scope and emphasis. The European Union's General Data Protection Regulation (GDPR) remains the archetype of a rights-based data framework, offering expansive protections including the `right to erasure' and strict processing limitations on personal data \cite{GDPR2016}. These provisions collectively enforce L1–L4 protections robustly and are often interpreted to cover AI training data. The EU's 2024 AI Act further builds on this by banning certain high-risk AI uses (L1), requesting data desensitization, requiring dataset documentation and labeling (L3), and reaffirming user-centric rights that overlap with L4 \cite{EUAIAct2024}; China's approach, through the Personal Information Protection Law (PIPL), Data Security Law, and the 2023 Measures for Generative AI, emphasizes state-centric oversight. These policies prohibit certain data uses (L1), require user consent and data anonymization (L2), and impose content labeling and prompt logging obligations (L3) \cite{PIPLPRC2021}. While deletability (L4) is nominally protected under Chinese law, enforcement practice remains limited; In contrast, the United States currently lacks a comprehensive federal data protection regime. Instead, privacy and deletion rights derive from sectoral and state-level statutes such as the California Consumer Privacy Act (CCPA), which supports L2 and L4 protections \cite{CCPA2018}. The Federal Zero Trust Data Security Guide reflects growing interest in traceability and risk-based governance (L3) but leaves implementation largely voluntary or agency-led \cite{federal2024}. In terms of regulatory design, the EU favors detailed, enforceable rights; China emphasizes preemptive control and compliance through licensing and supervision; and the U.S. leans on ex-post accountability and corporate commitments. Despite these differences, there is a broad convergence around the necessity of data non-usability, privacy-preservation, traceability, and deletability introduced by this paper.

Nonetheless, current regulations remain incomplete. A first major gap concerns \textit{cross-border enforceability}: data used in AI training often travels internationally, and fragmented legal standards create blind spots. For example, a dataset scraped in the U.S. and hosted in Singapore might be trained into a model deployed in Europe—yet only partial protections may apply depending on jurisdiction. Without global interoperability, enforcement becomes inconsistent and rights unevenly distributed \cite{oecd2025enhancing}. Second, even where rights to deletion exist (\eg, GDPR), technical feasibility lags. Removing data from AI models remains challenging once it has influenced model parameters (known as `model unlearning') \cite{bourtoule2021machine}. Regulatory texts rarely specify how such deletion should occur, leaving ambiguity in both compliance and remedy. Finally, many data protection laws remain focused on personal data. Large portions of AI training data involve non-personal but sensitive content: copyrighted content, synthetic datasets, or proprietary corpora. These fall outside privacy statutes and are instead covered unevenly under IP law or trade secret frameworks. Similarly, models themselves—containing learned representations of training data—are not clearly governed. Moving forward, regulators may need to embrace AI-specific rules for \textit{traceability by design} (\eg, mandatory dataset disclosure, logging, and watermarking), technical mandates for deletability, and broader coverage of non-personal data (\eg, artworks and models). Cross-border frameworks, such as global AI governance compacts or aligned certification standards, could help fill the compliance vacuum. In essence, while current regulations lay important foundations, future ones must evolve alongside the AI model's capabilities—embedding safeguards at every level of our taxonomy to ensure responsible innovation in the (generative) AI era.

\section{Discussions}
\label{sec:discussion}

While our proposed hierarchical taxonomy for data protection provides a critical guideline for the (generative) AI era, its establishment is a starting point, not an endpoint. This section moves beyond this foundational structure to dissect compelling emergent issues and underlying complexities that demand deeper exploration. Our goal is to spark the vital conversations needed to ensure data is handled responsibly and ethically as AI techniques and applications continue to evolve.

\vspace{-0.5em}
\subsection{Data Protection vs. Data Safety}
\label{subsec:protection_vs_safety}

Distinguishing between data protection and data safety is crucial, yet often overlooked. Data protection (in the AI era), as conceptualized in this paper through the hierarchy of non-usability, privacy-preservation, traceability, and deletability, fundamentally concerns the \textit{governance} and \textit{control} over data as an asset throughout the AI model's lifecycle \cite{li2023black,shao2025fit}. It addresses questions of ownership, authorized use, provenance, and the right to be forgotten --- essentially controlling how data flows and is utilized within the AI ecosystem, irrespective of the specific harm its content might cause. It focuses on safeguarding the rights and interests tied to the data itself and the models derived from it.

Data safety, in contrast, is primarily concerned with the \textit{content} of the data and the potential harms arising from that content or the model's behavior influenced by it \cite{liang2022advances, li2024safegen}. This includes issues like misinformation and deepfakes generated by models \cite{wang2024deepfake, li2024safeear}, biases encoded in training data leading to discriminatory outcomes \cite{mehrabi2021survey}, the generation of harmful or incorrect hazardous content \cite{zhang2025benchmark}, and the overall robustness and reliability of AI models and systems against adversarial manipulation aimed at causing malfunction or harm \cite{xie2023defending}. In essence, data safety seeks to \textit{mitigate the negative consequences} stemming from the data's substance or the AI model's outputs.

However, in the (generative) AI era, the lines between data protection and data safety are increasingly blurred and intertwined. Firstly, a lapse in one dimension often precipitates a failure in the other. For instance, a data-poisoning attack—classically a safety issue—can coerce a model into revealing sensitive training samples, thereby breaching privacy protections \cite{carlini2021extracting}; conversely, theft of a proprietary model—a protection failure—gives adversaries the means to mass-produce deepfakes or targeted misinformation, elevating safety risks \cite{li2025move,chen2025computational}; Besides, many countermeasures serve dual roles: watermarking, conceived as a traceability tool for data protection \cite{jia2020adv,liu2022watermark,ren2024sok}, also helps attribute and filter AI-generated misinformation, while access-control mechanisms, designed to safeguard data integrity, likewise prevent unauthorized generation of harmful content; More broadly, guarantees of data protection feed data-centric AI developing pipelines, improving dataset quality and control, thereby reducing bias, hallucination, \emph{etc.}—core challenges in data safety.

As we mentioned before, data safety, encompassing fairness, robustness, bias mitigation, and content moderation, is an equally critical but vast research area deserving its own dedicated treatment \cite{liang2022advances,mehrabi2021survey}. However, this paper concentrates primarily on the data protection dimension – establishing control over data assets within the AI lifecycle. In general, we focus on protection because establishing fundamental controls over data usage and provenance is often a prerequisite for tackling complex safety issues effectively. Nonetheless, recognizing the deep interplay is essential. Robust data protection mechanisms, particularly those ensuring traceability and controlled access, provide the foundational transparency and oversight needed to audit systems for safety concerns, attribute harmful outputs, and enforce safety-related policies. Future frameworks must holistically consider both aspects to build truly trustworthy (generative) AI models and systems.

\subsection{Emerging Challenges brought by AIGC in Data Protection}
\label{subsec:copyright_aigc}

The rise of AI-generated content (AIGC) powered by generative models introduces profound new challenges in data protection. In particular, many existing legal systems, including those in the US and EU, struggle to grant copyright protection to purely AIGC because it often lacks the requisite human authorship \cite{samuelson2023generative}. This leaves the ownership and copyrights associated with vast amounts of potentially valuable AIGC in a state of ambiguity. Who owns the novel image created by a diffusion model, or the code snippet generated by an LLM?

Rather than treating AIGC purely as content itself, our model-centric data protection perspective highlights further complexities. When AIGC is itself used as data – for instance, synthetic data for training new models, knowledge distillation \cite{hinton2015distilling}, or as input for retrieval-augmented generation systems – its copyright status becomes even more convoluted. Does the copyright (or lack thereof) of the original data used to train the generative model influence the status of the synthetic data? If a model distills knowledge from copyrighted data, does the resulting trained model (as a compact representation of information contained in these data) or the data it generates inherit restrictions? This debate touches upon the core definition of data rights: Are they solely tied to the direct expression of content, or do they extend to the statistical patterns, styles, and knowledge implicitly captured and transferable by a model \cite{shan2023glaze}? The potential for AI models (especially generative ones) to launder copyrighted information into seemingly novel, unprotected AIGC is a significant concern.

Even amidst this legal uncertainty, our proposed data protection framework offers valuable tools. The L3 (Traceability), through techniques like watermarking or fingerprinting \cite{ren2024sok, yao2024promptcare}, can help establish the provenance of AIGC, potentially linking it back to specific models or even training datasets. This provides crucial evidence for detecting plagiarism or unauthorized use of protected styles or content, even if the AIGC itself isn't copyrightable \cite{li2025towards}. Furthermore, L1 (Non-usability) techniques, such as data cloaking methods designed to disrupt style mimicry \cite{shan2023glaze}, offer technical safeguards for creators where legal protections are currently inadequate. These techniques and tools allow stakeholders to exert a degree of control over how their data or derived AI models influence future generations, shifting focus from solely legal ownership to technical prevention of undesired use.

Ultimately, these complex questions surrounding AIGC and copyright require urgent attention from policymakers and legal scholars. Future legislation must clarify the status of AIGC, define the boundaries of rights associated with training data and model-derived knowledge, and establish clear rules for the use and attribution of generated content. A specific protection framework like ours can inform these developments by highlighting what we need to protect and even what is technically feasible in terms of control and transparency.

\subsection{Challenges of Cross-Jurisdictional Data Protection}
\label{subsec:cross_jurisdictional}

The inherently global nature of the AI ecosystem presents significant hurdles for consistent data protection. The lifecycle of AI models, from data collection via web scraping or distributed sensors, annotation by global crowdsourcing platforms, training on cloud servers located potentially anywhere, to deployment for a worldwide user base, routinely cross multiple national borders. This immediately runs into the fragmented and vague landscape of international data protection regulations.

Currently, different jurisdictions have markedly different manners. The European Union's GDPR \cite{GDPR2016} imposes strict, rights-based obligations with extra-territorial reach. The US employs a sectoral approach supplemented by state-level laws like the CCPA \cite{CCPA2018}. China's PIPL \cite{PIPLPRC2021} emphasizes state oversight and data localization requirements. Other regions may have nascent or less comprehensive regulations \cite{oecd2025enhancing}. This regulatory patchwork creates significant compliance challenges for developers and opens avenues for exploitation. For example, data scraped in a jurisdiction with lenient rules might be used to train an AI model deployed in a region with strict privacy laws, creating legal jeopardy. Conversely, malicious actors might deliberately host AI models trained on improperly acquired data in jurisdictions with weak enforcement capabilities, undermining protection efforts globally.

Addressing these cross-jurisdictional challenges requires multifaceted solutions. On the policy front, greater international cooperation towards regulatory harmonization or establishing common minimum standards (perhaps through bodies like the OECD or UN initiatives) is desirable, although politically complex \cite{oecd2025enhancing}. Interoperability frameworks that allow different regulatory systems to recognize and interact with each other could offer a more pragmatic path than full unification. From a technical perspective, one approach is to adopt the strictest standard (\eg, GDPR compliance) globally, but this often imposes excessive costs and sacrifices utility unnecessarily in many contexts. A more promising direction lies in developing \textit{adaptive} data protection techniques. Future systems could potentially leverage context-aware mechanisms, perhaps inspired by meta-learning or zero-shot adaptation principles \cite{finn2017model}, to dynamically adjust protection levels (\eg, the type of watermarking, the rigor of privacy mechanisms, the implementation of deletion) based on the legal requirements of the data's origin, the user's location, or the operational jurisdiction. However, realizing such adaptive systems effectively still requires clear regulatory signaling and international collaboration on technical standards. Arguably, our hierarchical taxonomy can serve as a foundational conceptual framework – a common language – to facilitate these multi-stakeholder discussions, allowing different jurisdictions to map their specific requirements onto shared levels of protection, thereby aiding both policy alignment and the development of interoperable technical solutions.

\subsection{Ethical Considerations in Data Protection}

Beyond the conceptual and technical mechanisms and legal mandates, data protection in the AI era is intrinsically linked to fundamental ethical considerations. The choices made about how data is collected, used, shared, and managed reflect underlying values and have direct consequences for individuals and society. Our framework, while presented conceptually and technologically, implicitly engages with core ethical principles that warrant explicit discussion.

Arguably, the principle of \emph{autonomy} is central to this problem. Data privacy-preservation (\ie, Level 2) and data deletability (\ie, Level 4) directly support an individual's right to control their personal information and digital footprint, aligning with the `right to be forgotten' \cite{bourtoule2021machine}. Ensuring users have agency over their data is not just a legal requirement but an ethical imperative in an increasingly data-driven world. \textit{Fairness} is another critical dimension. While often discussed under data safety (\eg, mitigating algorithmic bias \cite{mehrabi2021survey}), protection mechanisms contribute significantly. Traceability (Level 3) enables audits to uncover biased data sourcing or discriminatory model behavior, fostering accountability. Preventing the unauthorized use of data (\ie, Level 1) can stop the malicious exploitation of vulnerable groups' data.

\textit{Transparency} and \textit{accountability} are cornerstones of ethical AI, directly supported by traceability. Knowing the provenance of data and models allows stakeholders to understand how systems work, assign responsibility for outcomes, and build trust. This is vital not only for redress but also for enabling informed public discourse about AI's role. Furthermore, the principles of \textit{beneficence} (doing good) and \textit{non-maleficence} (avoiding harm) are pertinent. Data protection helps ensure that the benefits of AI are realized responsibly. By preventing unauthorized access and misuse, it safeguards individuals from potential harms like identity theft, reputational damage from deepfakes, or the exploitation of creative work.

Navigating these ethical considerations often involves balancing competing values. There can be tension between maximizing data utility for societal benefit (\eg, in medical research) and upholding individual privacy. Innovation fueled by large datasets may clash with the rights of original data creators. The proposed hierarchy helps to make these trade-offs explicit, offering different levels of control to strike varying balances based on context and societal values. Responsibility for ethical data protection is shared across the entire AI lifecycle, involving data collectors, annotators, model developers, platform providers, deployers, and end-users. It requires fostering a culture of data stewardship that goes beyond mere legal compliance, embedding ethical reflection into the design, development, and deployment process. Our framework aims not only to provide conceptual, technical, and regulatory clarity but also to serve as a guideline and tool that encourages developers and policymakers to engage proactively with the profound ethical dimensions of data protection in the AI era.

\newpage
\section*{Acknowledgements}
We sincerely thank Prof. Bo Li (University of Illinois Urbana-Champaign) for her incisive insights and constructive suggestions from a professional side. We are also grateful to Prof. Dong Chen (Michigan State University), Jing Lyu (Columbia University), Yisheng Lin (Peking University), and Yiqiu Zhang (Shanghai AI Laboratory) for suggestions that broadened the relevance to a wider scientific audience. Finally, we acknowledge Chenfei Yao, Hua Tu, and Boheng Li (Nanyang Technological University) for their invaluable assistance in polishing Figures \ref{fig:intro}–\ref{fig:technique}, respectively.

\bibliographystyle{IEEEtran}
{\footnotesize \bibliography{ref}}

\end{document}